# Generative Choreography using Deep Learning

**Luka Crnkovic-Friis**
Peltarion
luka@peltarion.com

**Louise Crnkovic-Friis**
The Lulu Art Group
louise@theluluartgroup.com

## Abstract

Recent advances in deep learning have enabled the extraction of high-level features from raw sensor data which has opened up new possibilities in many different fields, including computer generated choreography. In this paper we present a system *chor-rnn* for generating novel choreographic material in the nuanced choreographic language and style of an individual choreographer. It also shows promising results in producing a higher level compositional cohesion, rather than just generating sequences of movement. At the core of *chor-rnn* is a deep recurrent neural network trained on raw motion capture data and that can generate new dance sequences for a solo dancer. *Chor-rnn* can be used for collaborative human-machine choreography or as a creative catalyst, serving as inspiration for a choreographer.

## Introduction

Can a computer create meaningful choreographies? With its potential to expand and facilitate artistic expression, this question has been explored since the start of the computer age. To answer it, a good starting point is to identify the different levels that go into a choreographic work.

A choreography can be said to contain three basic levels of abstraction, *style* (the dynamic execution and expression of movement by the dancer), *syntax* (the choreographic language of the work and choreographer) and *semantics* (the overall meaning or theme that binds the work into a coherent unit) (Blacking & Kealiinohomoko, 1979). All three levels present unique practical and theoretical challenges to computer generated choreography.

As syntax is the easiest to formalize in the form of a notation system, it has been the logical starting point for creating generative choreography (Calvert, Wilke, Ryman, & Fox, 2005). However, unlike music or literature dance lacks a universally accepted notation system. Although systems, such as Benesh movement notation and Labanotation have been proposed they have not been universally adopted mostly because of their steep learning curve (Guest, 1998). They also cannot capture *style* - the nuanced dynamics of movement that emerges as a collaboration between choreographer and dancer (Blom & Chaplin, 1982). The alternative of building computational models from raw data (video, motion capture) is alluring as it contains much more information. It has however until recently not been feasible both because of combination of lack of computing power, algorithms and available data (LeCun, Bengio, & Hinton, 2015). With the advent of GPU powered deep learning that has changed and we can now start building entirely data driven end-to-end generative models that are capable of capturing both *style* and *syntax*. Furthermore, as deep neural networks are capable of extracting multiple layers of abstraction, they can begin to model the *semantic* level as well. In this paper we describe such a system, *chor-rnn* that we have developed and discuss related work, show how the raw data is collected and present a deep recurrent neural network model. Finally, we also detail the training and discuss results, possible use and future work.

## Related work

Earlier work in this field has included various programmatic approaches with parametrized skeleton systems (Noll, 2013) as well as using simplified movement models combined with genetic algorithms to explore the parameter space (Lapointe, 2005). Several systems have been developed as a combination of a visualization system with a choice of pre-defined movement material that could be sequenced into longer compositions by the choreographer. Fully autonomous sequence generation has mostly been limited to sequencing a combination of snippets of movement material. Several proposed systems have been interactive, requiring a choreographer to make a number of selections during the generation phase (Carlson, Schiphorst, & Pasquier, 2011). Artificial neural networks have been used in generative systems in the past (McCormick, 2015). They have however not involved deep learning and the neural network presented in this paper is using tens of millions of model parameters rather than thousands.

## Recording movement

A choreography is the purposeful arrangement of sequences of motion. The basic building block is the change of position in a 3D space (Maletic, 1987). Techniques for recording the movement of human body in space are called "motion capture" and while here are various technical solutions at the time of writing, the most simple to use and cost effective was the Microsoft Kinect v2 sensor (Berger et al., 2011).

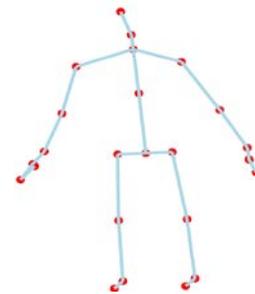

Figure 1 The red dots are joints tracked by the Kinect sensor.

It consists of a 3D camera augmented by an infrared camera and software that can automatically perform efficient and accurate joint tracking. The sensor records the movements of 25 joints (see Figure 1) at up to 30 frames per second. Each joint position is represented by a 3D coordinate for each frame. The sensor can in theory track up to 6 bodies simultaneously but multiple bodies can occlude each other relative to the field of view of the sensor. It has no way of tracking occluded joints (Fürntratt & Neuschmied, 2014). Multiple sensors can be used to overcome that limitation, but it requires more complex software to combine the results (Kwon et al., 2015). Our work was done with one sensor and one body.

## Generative model

Recurrent neural networks (RNNs) have been used to get state-of-the-art results for complex time series modeling tasks such as speech recognition and translation (Greff, Srivastava, Koutník, Steunebrink, & Schmidhuber, 2015). Since the motion capture data is a multidimensional time series we use a deep RNN model.

## Long Short-Term Memory

Standard RNNs are difficult to train in a stable way (due to the vanishing/exploding gradient problem) so we use a Long Short-Term Memory (LSTM) type of RNN. LSTMs are stable over long training runs and can be stacked to form deeper networks without loss of stability (Schmidhuber, 2015).
Contrary to a regular RNN which uses simple recurrent neurons, the central unit in an LSTM is a memory cell that holds a state over time, regulated by gates that control the signal flow in and out of the cell.
As the signal flow is tightly controlled, the risk is minimized of overloading the cell through positive feedback or extinguishing it through negative feedback.
The following equations show the relations in an LSTM cell (see Figure 2):

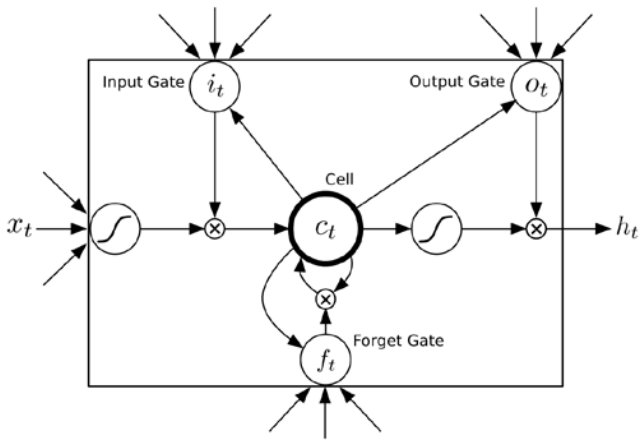

Figure 2 LSTM neuron

$$i_t = \delta(W^i x_t + R^i h_{t-1} + p^i \odot c_{t-1} + b^i) \quad (1)$$

$$f_t = \delta(W^f x_t + R^f h_{t-1} + p^f \odot c_{t-1} + b^f) \quad (2)$$

$$c_t = f_t \odot c_{t-1} \odot g(W^c x_t + R^c h_{t-1} + b^c) \quad (3)$$

$$o_t = \delta(W^o x_t + R^o h_{t-1} + p^i \odot c_{t-1} + b^i) \quad (4)$$

$$i_t = \delta(W_i x_t + R^i h_{t-1} + p^o \odot c_{t-1} + b^o) \quad (5)$$

Here $i_t$, $f_t$, $c_t$, $o_t$ and $h_t$ are the input gate, forget gate, memory cell and output gate at time step $t$; $x_t$ is the input while $\delta()$ and $g()$ are the sigmoid and tangent activation functions; $W$ and $R$ are the weight matrices applied to input and recurrent units; $p$ and $b$ are the peep-hole connection and biases while $\odot$ denotes dot product. Typically, when training a generative system, target output data would be the same as the input data but shifted with one sample:

$$t_t = x_{t+1} \quad (6)$$

This works well when the input and targets are discrete and the last layer is a softmax function (such as in the case of words or characters). For continuous functions as in this case there is a fundamental problem. When sampling dance movements 30 times/second $x_t$ is trivial to predict if $x_{t-1}$ and $x_{t-2}$ are known. It is just a continuation of the previous vector. A very simple model will produce very low errors during training, validation and testing:

$$x_t = x_{t-1} + (x_{t-1} - x_{t-2}) \quad (7)$$

However, when using it in a generative fashion where the output of the LSTM is used as the next input

$$y_t = LSTM_t(x_{t-1}) \quad (8)$$

it will fail completely. In cases where the data is discrete, a softmax introduces a controlled random element that can force the trajectory of the network into a new but controlled direction. In the case of continuous data, it is not possible as we do not have a controlled statistical distribution of the output so adding random noise will not help. In general, it can be shown that when using a mean square error metric, the output will stagnate and converge to an average output (Bishop, 1994).

## Mixture Density LSTMs

To counteract the issue of stagnating output we attach a mixture density network (MDN) to the output of the LSTM. This technique has been used successfully among other things for robotic arm control (Bishop, 1994) as well as handwriting generation (Graves, 2013).

Instead of just outputting a single position tensor, we output a probability density function for each dimension in the tensor. The output of the LSTM consists of a layer of linear output units that provide parameters for a mixture model defined as the probability of a target **t** given an input **x**:

$$p(\mathbf{t}|\mathbf{x}) = \sum_{i=1}^{m} \alpha_i(\mathbf{x})\varphi_i(\mathbf{t}|\mathbf{x}) \quad (9)$$

where $m$ is the number of components in the mixture with $\alpha_i$ being the mixing coefficients as a function of the inputs (**x**). The function $\varphi_i$ is the conditional density of the target tensor **t** for the i:th kernel. We use a Gaussian kernel, defined as:

$$\varphi_i(\mathbf{t}|\mathbf{x}) = \frac{1}{(2\pi)^{\frac{c}{2}}\sigma_i(\mathbf{x})^c} e^{-\frac{\|\mathbf{t}-\mu_i(\mathbf{x})\|^2}{2\sigma_i(\mathbf{x})^2}} \quad (10)$$

The neural network part that feeds into the mixture density model hence provides a set of values for the mixture coefficients, a set of means and a set of variances. The total number of output variables will be $m(c+2)$ where $m$ is the number of mixture components and c the number of output variables (a regular LSTM would have c outputs).

The outputs from the neural network will consist of a tensor

$$\mathbf{z} = [z_1^\alpha, \ldots, z_m^\alpha, z_{m+1}^\mu, \ldots, z_{mc+m+1}^\mu, z_{mc+m+2}^\sigma, \ldots, z_{m(c+2)}^\sigma] \quad (11)$$

containing all the necessary parameters to construct a mixture model. The number of mixture components, $m$, is arbitrary and can be interpreted as the number of different choices the network can pick at each time step. With the parametrized output, the whole MDN part can be encoded as a simple error metric where the error function becomes a negative log likelihood function (for the q:th sample):

$$E^q = -\log\left[\sum_{i=1}^{m} \alpha_i(\mathbf{x}^q)\varphi_i(\mathbf{t}^q|\mathbf{x}^q)\right] \quad (12)$$

Where

$$\alpha_i = \frac{e^{z_i^\alpha}}{\sum_{j=1}^{M} e^{z_j^\alpha}}, \quad \sigma_i = e^{z_i^\sigma}, \quad \mu_i = z_{ik}^\mu \quad (13)$$

and the derivatives needed for the training can be expressed as:

$$\frac{\partial E^q}{\partial z_k^\alpha} = \alpha_k - \frac{\alpha_k \varphi_k}{\sum_{j=1}^{m} \alpha_j \varphi_j} \quad (14)$$

$$\frac{\partial E^q}{\partial z_{ik}^\mu} = \frac{\alpha_i \varphi_i}{\sum_{j=1}^{m} \alpha_j \varphi_j} \frac{\mu_{ik} - t_k}{\sigma_i^2} \quad (15)$$

$$\frac{\partial E^q}{\partial z_i^\sigma} = -\frac{\alpha_i \varphi_i}{\sum_{j=1}^{m} \alpha_j \varphi_j} \left\{\frac{\|t - \mu_i\|^2}{\sigma_i^2} - c\right\} \quad (16)$$

The derivatives of the error function can be used with any standard gradient based optimization algorithm together with backpropagation.

**Training**

The data collected consisted of five hours of contemporary dance motion capture material created and performed by a choreographer. The resulting data set consisted of 13.5 million spatiotemporal joint positions. We used multiple deep configurations but the final neural network topology consists of 3 hidden layers with 1024 neurons in each (a total of ~21M weights). The input data was a 75-dimensional tensor (25 joints x 3 dimensions).

The model was trained for ~48h on a GPU computation server with 4 x Nvida Titan X GPUs (a total of 27 teraflops capacity). A batch size of 512 sequence parts (128/GPU) was used with a sequence length of 1024 samples. The sequence length corresponds to how many steps the system is unrolled in time and in effect the number of layers becomes 1024*3 = 3072 during training.

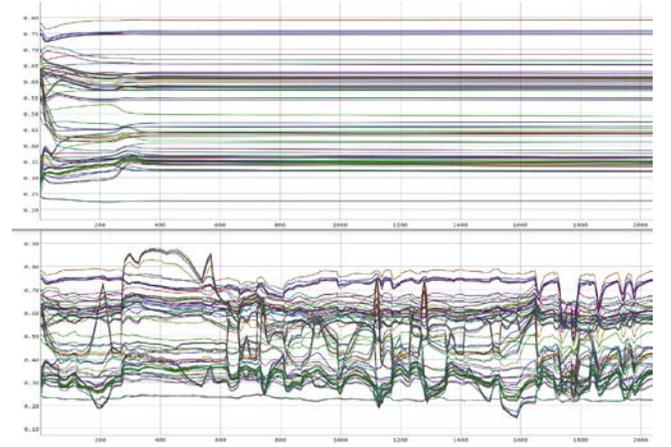

Figure 3 Output of a minute of generated joint positions over time: Without mixture density (top) and with mixture density (bottom)

The number of layers, and their effect in a recurrent neural network requires a far more complex interpretation than standard feed forward/convolutional neural networks as a signal can take an indeterminate number of spatiotemporal paths through the network. (Greff et al., 2015)

The neural network was trained with RMS Prop using Back Propagation Through-Time. The software was implemented in lua/Torch7 using the Peltarion Cortex platform. A comparison of a network trained with MDN and without can be seen in Figure 3.

In generation mode the MDN distributions were sampled at each time step to get a new set of coordinates for the joints. For this experiment we used unbiased sampling.

| Training Time | Sample frames from generated animation | | | Description |
|---|---|---|---|---|
| ~10 min | 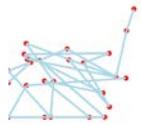 | 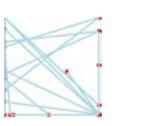 | 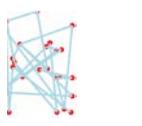 | Nearly untrained system. Joint positions are almost random.<br>https://www.youtube.com/watch?v=QnaKyc1Mpmo |
| ~6h | 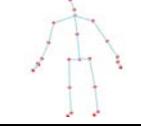 | 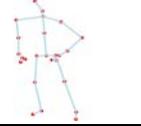 | 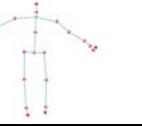 | Understands relative joint positions and very basic movement.<br>https://www.youtube.com/watch?v=c9h9zc7uPWQ |
| ~48h | 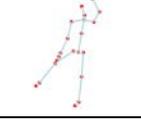 | 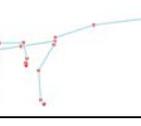 | 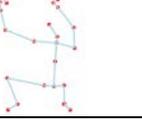 | Understands joint relations well, understand syntax and style well, understands basic semantics<br>https://www.youtube.com/watch?v=Q4_XSMqN8w0<br>https://www.youtube.com/watch?v=W1oRgDPxEkc |

Table 1 Example results over time

## Results

Our *chor-rnn* system can produce novel choreographies in the general style represented in the training data. Over the training interval it passes through several stages: basic joint relations (understanding the anatomy of human joints), basic movement style and syntax and at last the composition of several movements into a meaningful composition (semantics). See Table 1 for examples of results and Figure 4 for a visualization of example generated trajectories.

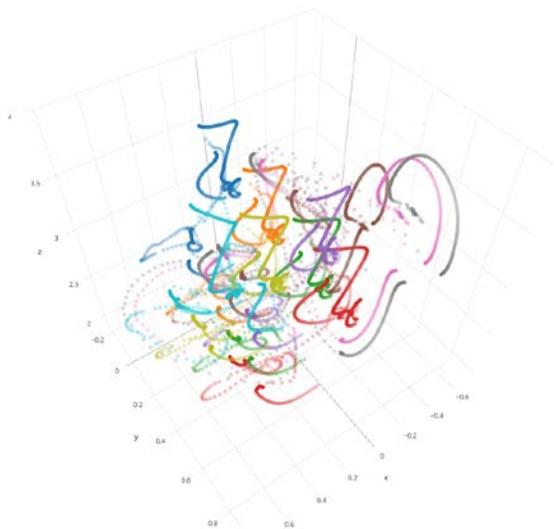

Figure 4 Spatial visualization of 30 seconds of generated trajectories for 10 joints. Each color represents a joint.

## Discussion

The generated material, presented as an animated "stick figure", was evaluated qualitatively by the choreographer. As choreography is an art largely based on physical expression and embodied knowledge (Blom & Chaplin, 1982), the choreographer also learned and executed the generated material. The conclusion was that the *chor-rnn* system produces novel, anatomically feasible movements in the specific choreographic language of the choreographer whose work it was trained on. If you generated an hours' worth of new choreography, it would have significantly less semantic meaning than the work it was trained on.

Generally speaking, the semantic level is the most difficult to quantify, especially when it comes to avant-garde art as it does not follow an established form (Foster, 1986). It is also the most complex one from the point of view of the neural network.

As with text or image generation (Hinton, 2014), the semantic level is the last one to emerge from the training.

There will of course be significant limitations when it comes to generating novel semantic levels as an artificial neural network can't draw on the human choreographer's life experience.

## Use as an artist's tool

While there are interesting philosophical questions regarding machine creativity especially in a longer perspective, it is also interesting to see how current results can be used as a practical tool for a working choreographer. The *chor-rnn* system in its current state can be used to facilitate a human choreographer's creative process in several ways. Two examples are collaborative choreography and mutually interpreted choreography.

In the first case the artist and *chor-rnn* can collaborate in creating a sequence by alternating between them as shown in Figure 5.

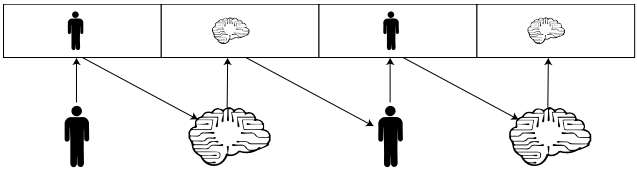

Figure 5 Alternating artist /chor-rnn choreography

1. The artist choreographs a sequence $A_1$
2. Chor-rnn takes sequence $A_1$ as input and produces a new sequence $B_1$ as a continuation of $A_1$
3. The artist looks at sequence $B_1$ and choreographs a new sequence $A_2$ as a continuation of $B_1$
4. Steps 2-3 are repeated

The resulting sequence will be $A_1B_1A_2B_2..A_NB_N$ – an alternation between human and computer choreography.

In the second case, the artist can start a sequence, let the *chor-rnn* generate a new sequence. The human can then reinterpret the output of the *chor-rnn* and input the interpretation into the system.

1. The artist choreographs a sequence $A_1$
2. Chor-rnn takes sequence $A_1$ as input and produces a new sequence (or set of sequences) $B_1$ as a continuation of $A_1$
3. The artist looks at $B_1$ and choreographs a reinterpretation of B1 as a new sequence, $A_2$
4. Steps 2-3 are repeated

The resulting sequence will be $A_1A_2A_3..A_N$ – a computer inspired human made choreography. Due to the symmetry of the process, a secondary sequence is created as well, $B_1B_2B_3..B_N$ – a human inspired computer made choreography.

When a choreographer works with a dancer to develop a choreography, the latter will inevitably influence the end result. While this may be desirable, it also dilutes the distinctive style (and possibly syntax) that is unique to the choreographer.

With *chor-rnn*, the choreographer works with a virtual companion using the same choreographic language. At the same time as it is capable of producing novel work, it can provide creative inspiration. As the level of machine involvement is variable and can be chosen by the choreographer, the results can be an interesting starting point of philosophical discussions on authenticity and computer generated art.

## Future work

**Collect a larger corpus of data** The five hours of motion capture data was enough to build a proof of concept system but ideally the corpus should be larger – especially if multiple choreographers are involved. For comparison state of the art speech recognition models use 100+ hours of data (and it is considered to be a major bottleneck in that field of research) (Graves & Jaitly, 2014).

**Derive a choreographic symbolic language** One of the most intriguing features of deep neural networks is that they internally build up multiple levels of abstraction (Hinton, 2014). Using a recurrent variational autoencoder would allow us to compress meaningful higher order information into a fixed size tensor (encoding) (Sutskever, Vinyals, & Le, 2014). This in turn would allow a derivation of a symbolic language and by mapping it to feature detectors that operate on that encoding.
A general symbolic encoding could provide an alternative to the existing notation systems and simplify the creation of computer created choreography. It could also provide a convenient method of recording a choreographic work in a compact, human readable format. As multiple mobile phone makers are now integrating 3D cameras (comparable to the Kinect) into their devices, an easy way of transforming recorded material to a symbolic encoding may be of significant practical use for documentation/archiving purposes (Kadambi, Bhandari, & Raskar, 2014).

**Multiple bodies** The Kinect sensor cannot directly handle occluded body parts. This is problematic even with one dancer and makes it nearly impossible to capture interactions between multiple dancers. The solution is to use multiple Kinect sensors and combine their data (Kwon et al., 2015). This would allow us to record choreographies with up to 6 dancers and allow the system to learn about interactions between dancers.

**Multi-modal input** The input data could be extended to beyond motion capture data also include sound (and even images and video). One could for instance build a system that in the generated choreography relates to a musical composition.

## Conclusions

This paper details a system, *chor-rnn* that is trained using a corpus of motion captured contemporary dance. The system can produce novel choreographic sequences in the choreographic style represented in the corpus. Using a deep recurrent neural network, it is capable of understanding and generating choreography *style*, *syntax* and to some extent *semantics*. Although it is currently limited to generating choreographies for a solo dancer there are a number of interesting paths to explore for future work. This includes the possibility of tracking multiple dancers and experimenting with variational autoencoders that would allow the automatic construction of a symbolic language for movement that goes beyond simple syntax. Apart from fully autonomous operation, *chor-rnn* can be used by a choreographer as a creativity catalyst or choreographic partner.

We asked if a computer could create meaningful choreographies and with tools like *chor-rnn* we think we can get one step closer to answering that question or at least to discover new relevant questions.